\documentclass[11pt]{article} %
\usepackage[preprint]{mldraft}

\usepackage[utf8]{inputenc} %
\usepackage[T1]{fontenc}    %
\definecolor{MyGreen}{HTML}{009900}
\definecolor{MyBlue}{HTML}{0d3983}
\RequirePackage[colorlinks,citecolor=MyGreen]{hyperref}
\usepackage{url}            %
\usepackage{booktabs}       %
\usepackage{amsfonts}       %
\usepackage{nicefrac}       %
\usepackage{microtype}      %

\usepackage{paperpkg}
\usepackage{subcaption}
\usepackage{textcomp}
\usepackage{soul}
\usepackage{pifont}

\RequirePackage{algorithm}
\usepackage{etoolbox}

\let\classAND\AND
\let\AND\relax
\usepackage{algorithmic}

\let\AND\classAND
\AtBeginEnvironment{algorithmic}{\let\AND\algoAND}

\usepackage[algo2e,ruled,vlined,linesnumbered]{algorithm2e}
\usepackage{pifont}
\usepackage{booktabs}
\usepackage{upquote}
\usepackage{listings}
\usepackage{comment}
\usepackage{fancyvrb}
\usepackage{graphicx}
\usepackage{multicol}
\usepackage{multirow}
\usepackage{cleveref}
\usepackage{lstautogobble}  %
\usepackage{color}          %
\usepackage{zi4}            %
\usepackage[breakable,skins]{tcolorbox}
\usepackage{listings}
\usepackage{xcolor}
\usepackage{diagbox}
\usepackage{booktabs}
\usepackage{colortbl}
\usepackage{bbding}
\usepackage{graphicx}
\usepackage{rotating}
\usepackage{adjustbox}
\usepackage{lipsum}
\usepackage{enumitem}
\usepackage{framed}

\usepackage{listings}

\theoremstyle{plain}
\newtheorem{theorem}{Theorem}[section]

\newtheorem{lemma}[theorem]{Lemma}

\theoremstyle{definition}

\theoremstyle{remark}

\newcommand{\algoname}{\texttt{LeetDecoding}}

\newcommand{\mytitle}{LeetDecoding: A PyTorch Library for Exponentially Decaying Causal Linear Attention with CUDA Implementations}

\title{\mytitle}

\author{%
\centerline{\name Jiaping Wang$^{\dagger,\ddagger}$\thanks{~~Equal contribution.}~~~
\name Simiao Zhang$^{\dagger,\ddagger}$\footnotemark[1]~~~
\name Qiao-Chu He$^{\mathsection}$~~~
\name Yifan Chen$^{\dagger}$
} \\[1ex]
\centerline{$^\dagger$ \addr Computational Machine Intelligence Laboratory, Hong Kong Baptist University}
\centerline{$^\ddagger$ \addr Software Engineering Institute, East China Normal University}
\centerline{$^\mathsection$ \addr School of Business, Southern University of Science and Technology}
}

\begin{document}
\maketitle

\begin{abstract}

The machine learning and data science community has made significant while dispersive progress in accelerating transformer-based large language models (LLMs), and one promising approach is to replace the original causal attention in a generative pre-trained transformer (GPT) with \emph{exponentially decaying causal linear attention}.
In this paper, we present \algoname, which is the first Python package that provides a large set of computation routines for this fundamental operator.
The launch of \algoname~was motivated by the current lack of 
\ding{182}~clear understanding of the complexity regarding this operator,
\ding{183}~a comprehensive collection of existing computation methods (usually spread in seemingly unrelated fields), and
\ding{184}~CUDA implementations for fast inference on GPU.
\algoname's design is easy to integrate with existing linear-attention LLMs, and allows for researchers to benchmark and evaluate new computation methods for exponentially decaying causal linear attention.
The usage of \algoname~does not require any knowledge of GPU programming and the underlying complexity analysis, intentionally making \algoname~accessible to LLM practitioners.
The source code of \algoname~is provided at \href{https://github.com/Computational-Machine-Intelligence/efficient_linear_decoding}{this GitHub repository}, and users can simply install \algoname~by the command \texttt{pip install leet-decoding}.

\end{abstract}

\section{Introduction}

The transformer~\citep{vaswani2017attention} architecture, especially since the inception of ChatGPT~\citep{OpenAI22}, has revolutionized the business application of AI techniques~\citep{ caiado2023ai,li2024fusion, finance,yang2024adllmbenchmarkinglargelanguage} particularly through the development of large language models (LLMs). 
Its core component, the softmax attention, is adept at modeling local and long-range dependencies~\citep{lu2019understandingimprovingtransformermultiparticle,tian2022inductive,wu2023on} and supports parallelized training, but suffers from quadratic complexity in sequence length~\citep{yang2022analyzing}.
To address the efficiency issue, numerous efforts have been made to accelerate the transformers;
notably, the startup company Groq, focusing on improving inference efficiency, had raised a \$640M Series D round at a valuation of \$2.8 billion in 2024~\citep{Moss_2024}.

\emph{Linear attention} has emerged as a promising alternative, especially for encoder transformers equipped with \emph{bidirectional attention}, by replacing the exponential similarity function in attention with a simple dot product over key/query token vectors (c.f. the introduction in \Cref{sec:edcla}). 
For example, Performer~\citep{choromanski2020rethinking} and random feature attention~\citep{peng2021random} decompose the attention weight matrix into a product of two rectangular matrices, respectively consisting of learned linear features or random features~\citep{rahimi2007random} of the keys and queries. 
Skyformer~\citep{chen2021skyformer} employed the \nystrom method for efficient computation, while LARA~\citep{zheng2022linear} merged randomized attention with random feature attention to effectively reconstruct the attention weight matrix. 
Moreover, Linformer~\citep{wang2020linformer} applied Johnson--Lindenstrauss (JL) transforms~\citep{Johnson1984ExtensionsOL,Dzahinihashing23} to attain fast computation; 
Skeinformer~\citep{chen-etal-2022-sketching} generalized the JL transforms to randomized sketching and approximated bidirectional attention with sampling techniques in approximately linear time to improve inference speed. 
Transformer-VQ~\citep{lingle2023transformer} performed ``clustering'' on the sequence of Key vectors in attention using vector quantization, approximating the original vectors with the centroid of their respective classes. 

However, despite the linear scaling in memory / computational requirements and the empirical successes seen in encoder transformers, 
few linear-attention attempts are made to match the performance of \emph{decoder transformers} (i.e., GPT-like transformers), which are exampled by ChatGPT and thus the current mainstream paradigm for large language models.
Specifically, a decoder block in a transformer will adopt \textbf{causal attention} (i.e., unidirectional attention, c.f.\ \Cref{sec:standard-attn}), and a vanilla application of linear attention techniques to causal attention (namely \textbf{causal linear attention}) will still lead to an $\m O(N^2)$ complexity, as shown in \Cref{eq:BCMV}.
This plausible obstacle does hinder the exploration of adapting preceding linear attention methods to causal attention mechanisms;
while it turns out the intrinsic complexity for causal linear attention is $\m O(N)$ as well, most LLM practitioners are unaware of the analysis result.

Until recently, the \emph{exponential decay of attention scores} has been found useful to bridge the performance gap between causal attention and \textbf{causal linear attention},
which is now the de facto configuration taken by various \emph{linear transformers} (linear attention transformers trained from scratch).
For example, RWKV~\citep{peng-etal-2023-rwkv}, RetNet~\citep{sun2023retentive}, and TransNormerLLM~\citep{qin2023transnormerllm} combine causal masking and exponential decay along relative distance into a single matrix. 
This approach alleviates attention dilution issues while maintaining computational efficiency.
Notably, the linear-attention-based LLM, TransNormerLLM, with the design of exponential decay, has been reported to outperform conventional softmax-attention-based models in both accuracy and efficiency~\citep{qin2023transnormerllm};
RWKV, instead of the more prominent model GPT-4, has even been incorporated into the Windows 11 system~\citep{Rwkv_2024} due to its superior efficiency.

In spite of the fast development of linear transformers,
existing computation methods are spread in various fields/methods (summarized in \Cref{tab:calculating}) and are not yet realized by the community of machine learning and data science that they are appropriate to causal linear attention,
let alone available via a comprehensive, easy-to-use library.
Whereas there are recent studies focusing on optimizing regular linear attention on modern GPUs and give I/O-aware algorithms implementations~\citep{yang2023gated}, 
they only implement a specific calculation and does not consider and analyze scenarios under different batch sizes such as in \Cref{sec:empirical_results}. 
Moreover, the code of \citet{yang2023gated} is highly coupled and is not well compatible with other large models.
There is a strong need for a fast, user-friendly library to efficiently run transformers with exponentially decaying causal linear attention.

\begin{table}[t]
\caption{Overview of the computation methods for causal linear attention.}
\label{tab:calculating}
\centering
\scalebox{0.97}{
\begin{tabular*}{\textwidth}{cccc}
\hline 

\multicolumn{2}{c}{Computation Method} & Source Field & Time Complexity  \\
\hline 
\multicolumn{2}{c}{vanilla (\Cref{eq:BCMV})} & - & $\m O(N^2)$ \\
\multicolumn{2}{c}{causal-dot-product~\citep{vyas_et_al_2020}}  & linear attention & $\m O(N)$ \\
\multicolumn{2}{c}{block-based~\citep{lingle2023transformer}} & linear transformer & $\m O(N)$  \\
\multicolumn{2}{c}{recursion~\citep{han2023hyperattention}} & attention approximation & $\m O(N\log{N})$  \\
\multicolumn{2}{c}{lightningAttention-2~\citep{qin2024lightning}} & linear transformer & $\m O(N)$  \\
\multicolumn{2}{c}{FleetAttention (\Cref{sec:fleetAttention})} & this paper& $\m O(N)$  \\
\hline 
\end{tabular*}
}
\end{table}

\subsection{Our contributions}

In this work, we thoroughly investigate existing linear-complexity computation algorithms from various fields (not originally for causal linear attention), expanding and adapting these algorithms for standard low-rank attention. 
To the best of our knowledge, our work is the first study in this area. 
Additionally, we provide complexity analyses for multiple computation methods, which are usually dismissed by LLM practitioners.
In particular, one recursive computation method~\citep{han2023hyperattention} indeed has a suboptimal $\m O(N \log N)$ complexity (c.f.\ \Cref{sec:recursion}) and thus the practical efficiency is inferior to other candidate methods, as shown in \Cref{sec:empirical_results}.

More importantly, we introduce \algoname, a Python package that adapts and implements all the preceding computation methods for exponentially decaying causal linear attention, along with a newly proposed algorithm FleetAttention. 
Furthermore, the package is built on PyTorch with CUDA optimizations to leverage the parallel processing capabilities of modern GPUs. 
We demonstrate that \algoname~is easy to use for individual decoder modules and can be seamlessly incorporated into large language models. 
These two features are especially useful to LLM practitioners as, in deploying LLMs, intensive performance evaluations are a must to empirically decide the optimal implementations under different scenarios. (See the results in \Cref{sec:empirical_results}. No computation method can dominate others along all the settings.)

\subsection{Paper organization}

In the remainder of this paper, \Cref{sec:notation} introduces necessary notations and preliminaries of exponentially decaying causal linear attention. 
In \Cref{sec:library_design}, we introduce the algorithms involved in the package and the CUDA implementation for fast inference on GPU. 
In \Cref{sec:how_to_use}, we demonstrate the usage of the package \algoname, which supports both benchmarking computation methods in individual causal linear attention modules and integrating with existing large linear transformers. 
Empirical results are provided in \Cref{sec:empirical_results} for both of the settings above.
We leave a concluding remark in \Cref{sec:conclude}.

\section{Notations and Preliminaries}
\label{sec:notation}

For the reader's convenience, we introduce the standard attention mechanism as a preliminary in \Cref{sec:standard-attn}.
Exponentially decaying causal linear attention, the core of this library, is then introduced in \Cref{sec:edcla}.

\subsection{Standard attention mechanisms}
\label{sec:standard-attn}

The dot-product attention~\citep{vaswani2017attention} processes three input matrices $\mtx{Q},\mtx{K},\mtx{V} \in \mathbb{R}^{N \times d}$, wherein $N$ is the number of tokens in the input sequence and $d$ denotes the head dimension. 
On one hand, the bidirectional attention, the first-generation attention proposed in the BERT language model~\citep{devlin2018bert}, is defined as:
\begin{equation}
\label{equ:bidirec}
\mtx{D}^{-1}\mtx{A}\mtx{V}
\end{equation}
Here the matrix $\mtx{A}=\exp(\mtx{Q}\mtx{K}^T) \in \mb{R}^{N \times N}$ represents the element-wise exponential of $\mtx{Q}\mtx{K}^T$ (for simplicity, the normalization factor $\sqrt{d}$ has been omitted). 
The diagonal matrix $\mtx{D}=\mathrm{diag}(\mtx{A} \mtx 1_N)$ contains the row sums of $\mtx{A}$ along its main diagonal, where $\mtx 1_N$ is a length-$N$ all-ones vector. In this context, matrix $\mtx{A}$ is called the \textit{attention matrix} or \emph{attention score matrix}.

On the other hand, causal attention (or unidirectional attention) is used for auto-regressive generative modeling, such as the decoder components of ChatGPT and its successor GPT-4~\citep{OpenAI22}. 
It is defined as:
\begin{equation}
\begin{aligned}
\label{equ:unidirec}
(\mtx{D'})^{-1} (\mtx{A} \odot \mtx{M}) \mtx{V},
\quad \text{where~}
\mtx{D'} &= \mathrm{diag}((\mtx{A} \odot \mtx{M}) \mtx 1_N),
\end{aligned}
\end{equation}
and $\mtx{M} \in \{0,1\}^{N \times N}$ is a lower triangular matrix:
$$\mtx{M}_{i,j} = \delta(i, j) \defeq
\begin{cases} 
1, & \text{if } i \geq j, \\
0, & \text{otherwise}.
\end{cases}
$$
The time complexity for computing both \Cref{equ:bidirec} and \Cref{equ:unidirec} is $\m O(N^2d)$. Their space complexity is $\m O(N^2+Nd)$, which is due to the explicit use of the attention matrix $\mtx{A}$.

\subsection{(Exponentially decaying) Causal linear attention}
\label{sec:edcla}

Researchers have invented various linear-attention large language models to mitigate the quadratic computational complexity associated with standard attention mechanisms, including TransNormerLLM~\citep{qin2023transnormerllm}, RetNet~\citep{sun2023retentive}, RWKV~\citep{peng-etal-2023-rwkv}, and Mamba~\citep{gu2023mamba}. 
The computation of attention matrix in these linear attention models can all be abstracted into a low-rank form involving matrices $\mtx B, \mtx{C}\in \mathbb{R}^{N\times d}$ where we formally remove the element-wise exponential function and the row normalization in \Cref{equ:bidirec}. 
A universal formula,
\begin{equation} \label{eq:universal}
\mtx{A} = \mtx{B}\mtx{C}^T,
\end{equation}
can succinctly describe the attention matrix functionality within these models and is used along this paper.
For TransNormerLLM and RetNet, for example, $\mtx{B}=\mtx{Q}$, $\mtx{C}=\mtx{K}$.

With \Cref{eq:universal} at hand, we can formulate the \emph{exponentially decaying causal linear attention} as follows:
\begin{equation}
\label{eq:BCMV}
(\mtx{B}\mtx{C}^T\odot\mtx{M})\mtx{V}
\end{equation}
where the lower triangular matrix $\mtx M$ is now reloaded, with 
$$\mtx{M}_{i,j} = \delta(i, j) \defeq
\begin{cases}
\gamma^{i-j}, & \text{if } i \geq j, \\
0, & \text{otherwise}.
\end{cases}
$$
In the case $\gamma = 1$, \Cref{eq:BCMV} will be simply dubbed \emph{causal linear attention}.

In addition to the aforementioned linear attention adopted in a native linear transformer, all \emph{low-rank approximation methods} can be similarly unified in the form of \Cref{eq:universal}. 
Low-rank approximation methods, when applied to bidirectional attention mechanism, involve the utilization of two matrices $\mtx{B}, \mtx{C} \in \mathbb{R}^{N\times r}$ of low rank (here we reload the notations for arbitrarily two matrices $\mtx{B}, \mtx{C}$), to approximate the attention matrix $\mtx{A}$ through $\tilde{\mtx{A}} = \mtx{B}\mtx{C}^T$,
where $r$ indicates the rank of $\tilde{\mtx{A}}$ and conceptually differs from the previous (head) dimension $d$. 
We illustrate the ubiquity of the approximation framework above through dissecting Transformer-VQ~\citep{lingle2023transformer} as an example;
in their approximation method, $\mtx{B}=\mtx{Q}$ and $\mtx{C}=VQ(\mtx{K},\mtx{c})$ where $\mtx c$ is a codebook.

In applying those low-rank techniques for either linear attention or attention approximation, however, we formally still need to first obtain \Cref{eq:BCMV};

however, due to the application of a mask to~$\tilde{\mtx{A}}$, we cannot directly utilize the low-rankness thereof and reduce the computation complexity by matrix right-multiplication as in bidirectional attention. 
A na\"ive implementation of \Cref{eq:BCMV} still leads to a time complexity $\m O(N^2r+N^2d)$ and memory complexity $\m O(N^2+Nd)$.
Shortly in \Cref{sec:fleetAttention}, we will illustrate the intrinsic linear complexity of \Cref{eq:BCMV} through introducing a new \emph{computation method}, FleetAttention, which rearrange the computation process of \Cref{eq:BCMV} to attain linear complexity.

\section{Algorithms and Library Design}
\label{sec:library_design}

The overall goal of our library is to enhance the computational efficiency of the attention component in transformer models. 
In \Cref{sec:fleetAttention}, we introduce a novel computation method named FleetAttention for calculating causal linear attention and showcase its intrinsic linear complexity. 
\Cref{sec:othermethods} lists all the collected efficient methods (to the best of our knowledge) for computing linear attention that are implemented in our library. 
In \Cref{sec:recursion}, we analyze the asymptotic suboptimality of using recursive methods for computing linear attention. 
Finally, \Cref{sec:implementation} delves into the specific implementation details of the methods within our library.

\subsection{A non-iterative matrix formulation for exponentially decaying causal linear attention}
\label{sec:fleetAttention}

In this section, we introduce a new computation method named FleetAttention for computing exponentially decaying causal linear attention. Our method employs a non-iterative matrix formulation and thus clearly exhibits the intrinsic complexity of exponentially decaying causal linear attention.

Given the input matrices $\mtx{B}, \mtx{C}, \mtx{V}$, our objective is to compute the attention output $\mtx{O}$ with linear compute complexity. 
Here, we mainly describe how to compute $\mtx O \defeq (\mtx{B}\mtx{C}^T\odot\mtx{M})\mtx{V}$ in \Cref{eq:BCMV}. 
To ease the derivation, we denote $\mtx b_i, \mtx c_i$ respectively as the $i$-th column of $\mtx{B}, \mtx C$, and thus $\mtx{B}\mtx{C}^T$ amounts to $\sum_i \mtx b_i \mtx c_i^T$.
We then have:
\begin{align*}
\begin{split}
\mtx O &= \sum_{i=1}^r \paren{ (\mtx b_i \mtx c_i^T) \odot \mtx{M} }\mtx{V} = \sum_{i=1}^r \mathrm{diag}(\mtx b_i)\mtx{M} \mathrm{diag}(\mtx c_i)\mtx{V},
\end{split}  
\end{align*}  
in which the second equation holds due to the known property of Hadamard product. For a classic causal mask with binary elements in $\{0, 1\}$, we leverage the special structure of the mask matrix $\mtx M$ that it is equivalent to a $\mathrm{cumsum}$ (cumulative sum) operator,
we finally obtain
\begin{align}
{\mtx O} = \sum_{i=1}^r \mathrm{diag}(\mtx b_i)\mathrm{cumsum}(\mathrm{diag}(\mtx c_i) \mtx{V}), 
\label{equ:p}
\end{align}
where the $\mathrm{cumsum}$ operator is applied to the column vectors of matrix $\mathrm{diag}(\mtx c_i) \mtx{V}$ in parallel.
We note in obtaining each summand the time complexity is $\m O(Nd)$, and therefore the total computation complexity will remain $\m O(Ndr)$.
The algorithm above can be generalized to a popular attention variant with exponentially decaying positional weights~\citep{press2022train,zhen2022cosformer,peng-etal-2023-rwkv}, and the corresponding derivation is provided in \Cref{sec:discount_cumsum}.

The FleetAttention method has been incorporated into the library, as shown in \Cref{{sec:othermethods}}.
As a closing remark, although this method clearly showcases the intrinsic linear complexity of exponentially decaying causal linear attention, fine-grained CUDA programming cannot apply to its practical implementation as most of its calculation process is Hadamard product, rather than conventional matrix multiplication favored by GPU. 
Although operations along the rank dimension can be performed in parallel, the final process of summing up the results is mutually exclusive and must be carried out serially; detailed analysis is deferred to \Cref{sec:analysis_FleetAttention}.

\subsection{Implemented methods}
\label{sec:othermethods}

We unearth the algorithms for causal linear attention from the literature in various fields (we indeed generalize some methods from their original settings to low-rank causal attention), and prepare the PyTorch implementation if originally unavailable.
Table~\ref{tab:calculating} has already presented a comprehensive overview of existing computation algorithms.
We give a brief introduction to each algorithm in the following list; more detailed description, including how to generalize them from the original setting, are provided in \Cref{sec:other}.

\paragraph{\texttt{vanilla}.} The direct computation of \Cref{eq:BCMV} using PyTorch. We implemented the computation with both casual masks that employ binary values in $\{0, 1\}$ and masks with exponentially decaying positional embedding.

\paragraph{\texttt{causal-dot-product}.}
This algorithm was first introduced by \citet{vyas_et_al_2020} as the \texttt{CausalLinearAttention} class in their Fast Transformers library\footnote{See \url{https://github.com/idiap/fast-transformers}.} (while this algorithm was not detailed and explained in their paper). 
This algorithm utilizes the dot product between feature maps ($\mtx B, \mtx C$) and leverages the triangular structure of the causal masking to compute the attention outputs row by row, achieving linear time complexity. 
\citet{choromanski2020rethinking} reinvented the computation method \texttt{causal-dot-product} for addressing general low-rank formats, as the algorithm description in their appendix is consistent with the code of \texttt{causal-dot-product}.
Full details of this algorithm can be seen in \Cref{al:performer}.

The official CUDA implementation of \texttt{causal-dot-product} by \citet{vyas_et_al_2020} solely supports the fp32 type and the classic causal linear attention calculations with masks consisting exclusively of 0s and 1s on GPU. 
We modified its code to support both fp32 and fp16 types, as well as the causal masks with exponentially decaying positional embedding.
Moreover, as there is no native PyTorch implementation provided in either \citet{vyas_et_al_2020} or \citet{choromanski2020rethinking},
we develop a PyTorch version of \texttt{causal-dot-product}, i.e., \texttt{causal-dot-product\_torch}, for the library user's convenience.

\paragraph{\texttt{block-based}.} 
In the attention approximation method Transformer-VQ~\citep{lingle2023transformer}, they proposed an algorithm to accelerate causal linear attention computation through a block-by-block approach. 
The original design is uniquely tailored to the vector-quantized keys and initially applied in a gated activation unit (GAU). 
We have extended this algorithm to a more general low-rank factorization format as in \Cref{eq:BCMV}. 
The extended algorithm is detailed in Appendix~\ref{al:vq} and implemented using PyTorch. 

\paragraph{\texttt{recursion}.} 
\citet{han2023hyperattention} proposed to compute causal linear attention by dividing the attention matrix into masked and unmasked submatrices (c.f. Appendix~\ref{al:recur}); 
then, they utilized sparse attention to accelerate the computation of the unmasked submatrix, while continuing to recursively divide the masked submatrices. 
For the PyTorch implementation in \algoname, we substantially improve its computational efficiency by running \texttt{causal-dot-product} to calculate its base case.

\paragraph{\texttt{lightningAttention-2}.} 
\citet{qin2024lightning} leveraged the concept of ``divide and conquer'' by
separately handling the intra-block and the inter-block components in causal linear attention calculation. 
They divide a long input sequence into multiple blocks;
inside a block, they simply use \texttt{vanilla} to compute the product $(\mtx{B}\mtx{C}^T\odot\mtx{M})\mtx{V}$, while at the inter-block level they follow \texttt{causal-dot-product} and apply the similar linear attention kernel trick. 
More algorithm details can be seen in Section~3.2 of \citet{qin2024lightning}.

Its original code is implemented by OpenAI Triton while does not support the fp32 type, and even only supports NVIDIA A100 GPU;
the causal masks involved are required to have exponentially decaying positional embedding. 
To generalize their implementation for other scenarios, we modified its source code so that the subroutine can run on GPUs other than NVIDIA A100, allow fp32 types, and natively support regular causal masks with binary values. 
In addition, we also implemented this method in PyTorch and named it \texttt{lightningAttention-2\_torch} for more comprehensive evaluation.

\paragraph{\texttt{FleetAttention}.} 
We recall the method proposed in \Cref{sec:fleetAttention} adopts a non-iterative matrix formulation to compute causal linear attention; 
through leveraging the cumulative sum (\texttt{cumsum}) operation in \Cref{equ:p}, the computational complexity is reduced to $\m O(N)$. 

We used OpenAI Triton to write the core algorithm of FleetAttention and supported causal masks with both binary values in $\{0, 1\}$ and exponentially decaying positional embedding.
(The details of its GPU programming can be found in \Cref{app:FleetAttention}.)
Its native PyTorch implementation is provided in \algoname~as well and named \texttt{FleetAttention\_torch}.

\subsection{Asymptotic suboptimality of recursive computation method}\label{sec:recursion} 

An interesting observation of the lower triangular matrix $\mtx{M}$ is that the product $(\mtx{M} \odot \tilde{\mtx A}) \mtx V$ can be rewritten in a partitioned manner (for simplicity we omit the dependence on $\tilde{\mtx A}$ in the notations of $\mtx M_i$'s):
\begin{align}
\begin{pmatrix}
\mtx M_1 & \mtx 0 \\
\mtx M_2 & \mtx M_3
\end{pmatrix}
\begin{pmatrix}
\mtx V_1 \\
\mtx V_2
\end{pmatrix} = 
\begin{pmatrix}
\mtx M_1 \mtx V_1 \\
\mtx M_2 \mtx V_1 + \mtx M_3 \mtx V_2
\end{pmatrix},
\label{eqn:partition}
\end{align}
where $\mtx M_1, \mtx M_3$ are two smaller lower triangular matrices and $\mtx M_2$ in the bottom left corner of $\mtx{M} \odot \tilde{\mtx A}$ is a regular dense matrix.
Realizing the observation, \citet{han2023hyperattention} mentioned a scheme to efficiently compute their proposed low-rank attention in decoders: we can normally compute the product $\mtx M_2 \mtx V_1$ and then recursively address the two smaller matrices $\mtx M_1 \mtx V_1$ and $\mtx M_3 \mtx V_2$.

The recursive strategy stands distinct from other efficient causal linear attention computation methods. 
Although the strategy is attractive at first sight, we note it suffers from several deficiencies, \circled{1} higher time complexity and \circled{2} inefficient hardware implementation.

\circled{1} As shortly delineated in Lemma~\ref{lem:recur}, even equipped with an efficient $\m O(N)$ low-rank attention computation method, the recursive approach is characterized by a computational complexity of $\m O(N \log N)$, with its proof detailed in \Cref{sec:recur}.  
We note the recursive strategy exhibits a higher complexity than the original $\m O(N)$ low-rank attention computation method. 
\begin{lemma}
\label{lem:recur}
Consider the matrices $\mtx{B}$, $\mtx{C} \in \mb{R}^{N \times r}$, $\mtx{V} \in \mb{R}^{N \times d}$, $\mtx{\tilde{A}} = \mtx{B} \mtx{C}^T$, and the causal mask matrix $\mtx M$. 
The time complexity of using the recursive computation method to compute $(\tilde{\mtx{A}}\odot \mtx{M}) \mtx{V}$, even equipped with a linear-complexity algorithm to compute the base case, is $\m{O}(N\log N)$.
\end{lemma}
\circled{2} In addition, GPUs are designed to optimize large-scale, parallel data processing, while recursion involves deeply nested calls and complex control flows, which is inconsistent with the advantages of modern hardware. 
Combining \circled{1} and \circled{2}, we conclude the recursive computation method is sub-optimal for decoding extremely long prompts on GPUs.

\subsection{Fast inference on GPU and CUDA implementation}
\label{sec:implementation}

Implemented in PyTorch, all the methods in \Cref{sec:othermethods} can naturally run on GPU.
For most methods, the native PyTorch implementation is sufficient; however, for \texttt{causal-dot-product}, \texttt{lightningAttention-2}, and \texttt{FleetAttention}, lower-level GPU programming in languages such as native CUDA and OpenAI Triton can improve parallelism and better utilize GPU resources. 
To more efficiently leverage the computational power of GPUs, we heavily optimize the CUDA implementations of the three methods above,
and details can be found in \Cref{app:cuda-details}.

In \Cref{sec:single}, we exhibits the efficiency difference between the CUDA implementation and the native PyTorch implementation; the former in most cases is more than twice faster than the counterpart native Pytorch implementation.
\Cref{tab:methods_impl} summarizes the implementation status of the methods in \algoname, and remarkably all of them support both fp16 and fp32 types on GPU.

\begin{table}[t!]
\centering
\caption{The programming language used to implement the methods in the LeetDecoding library.
`-' indicates the PyTorch implementation is already sufficient and a CUDA implementation cannot further accelerate the method.
}
\begin{tabular}{cccc}
\hline
Method name & PyTorch impl.\ & CUDA impl.\ & Abbreviation\\ \hline
vanilla & \Checkmark~ & - & Vanilla \\ 
block-based & \Checkmark~ & - & Bb\\ 
causal-dot-product & \Checkmark~ & Native CUDA~ & Cdotp \\ 
FleetAttention & \Checkmark~ & OpenAI Triton~ & FA\\ 
lightningAttention-2 & \Checkmark~ & OpenAI Triton~ & LA\\ 
recursion & \Checkmark~ & - & Recur \\ 
\hline
\end{tabular}
\label{tab:methods_impl}
\end{table}

\section{How to Use \algoname}
\label{sec:how_to_use}

We introduce how \algoname~is designed to serve various LLM practitioners. At the end of this section, we as well provide a snippet to demonstrate the practical usage of \algoname.

\circled{1}~For application developers adopting linear transformers to achieve supreme inference efficiency, \algoname~can be integrated into transformer-based models through its API. 
For transformer-based models, you can directly replace the original attention computation interface with \algoname~for attention calculation.
The input format of this library is consistent with the most popular computation libraries like HuggingFace Transformers\footnote{See \url{https://github.com/huggingface/transformers}} or FlashAttention\footnote{See \url{https://github.com/Dao-AILab/flash-attention}}.

\begin{lstlisting}[caption={An example of how to use LeetDecoding Library to benchmark your own implementation.},label={lst:benchmark}, language=Python]
from efficient_linear_decoding.test_method import benchmark_method

def attn(q, k, v, gamma):
    # customize the linear attention computation method here.
    pass

# Example usage of the benchmark_method() function:
# Note: The values below should be replaced with the most suitable values.

benchmark_method(method=attn, device='cuda', dtype='float32', batch_size=16, is_weight_decay=True,gamma=gamma)
\end{lstlisting}

\circled{2}~For researchers in the community of machine learning and data science, \algoname~can serve as a benchmark for profiling the newly proposed linear attention algorithms, as shown in \Cref{lst:benchmark}. 
\algoname~is intentionally designed for easy extension, and new computation methods can be implemented therein through creating the corresponding method class and subsequently registering the class in \algoname.

To demonstrate the simplicity and the convenience of the coding style in \algoname, we provide another snippet to show how \algoname~solves synthetic problems efficiently by a few lines of code.  
Specifically, \Cref{lst:use_leetDecoding} gives three examples to illustrate how to use the \algoname~library, ranging from the passing of input arguments, to the specification of the mask type and the computation method.

\begin{lstlisting}[caption={Inference using the causal\_linear\_decoder requires one line for regular causal attention or two lines with exponentially decaying mask.}, label={lst:use_leetDecoding},language=Python]
from efficient_linear_decoding import causal_linear_decoder

batch_size,heads,seqlen,rank = 1,32,4096,128
dim = 256
B = torch.randn((batch_size,heads,seqlen,rank),device='cuda')
C = torch.randn((batch_size,heads,seqlen,rank),device='cuda')
V = torch.randn((batch_size,heads,seqlen,dim),device='cuda')
gamma = torch.full((heads,1),0.9,device='cuda')

# If the user intends to apply a 0-1 causal mask.
output = causal_linear_decoder(B,C,V)

# If the user intends to apply an exponentially decaying mask.
output = causal_linear_decoder(B,C,V,is_mask_weight=True,gamma=gamma)

# If the user intends to apply a specified method in the library.
output = causal_linear_decoder(B,C,V,attn_method='block-based')
\end{lstlisting}

\section{Empirical Results}
\label{sec:empirical_results}

We report the empirical results in this section.
In \Cref{sec:single}, we evaluate the computational performance of implementations in our library when handling \emph{extremely long prompts}. 
Then in \Cref{sec:exp-linear-transformer}, we evaluate the aforementioned computation methods under different settings on pretrained linear transformers. 
Moreover, along this section we adopt the abbreviations in \Cref{tab:methods_impl} to simplify the presentation.

\subsection{Decoding extremely long prompts}
\label{sec:single}

We investigated the efficiency on a single-layer (exponentially decaying) causal linear attention module for the methods implemented in our library, across various prompt lengths\footnote{This experimentation was carried out on a 48GB NVIDIA A6000 GPU.}. 

Inference latency was measured for prompt lengths ranging from 128 to 100,000 tokens. \Cref{tab:single_bs_1_wo_decay} and \Cref{tab:single_bs_1_with_decay} show results with and without the exponentially decaying mask, both using a batch size of 1. Similarly, \Cref{tab:single_bs_16_with_decay} and \Cref{tab:single_bs_16_wo_decay} show results with and without the exponentially decaying mask, both using a batch size of 16\footnote{Along all the tables in this paper, boldface represents the best result, and ``OOM'' denotes ``out-of-memory''.}. 

Each measurement was replicated 15 times, and the final results are reported as the mean value and the standard deviation of each configuration.

\begin{table}[t!]
\centering

\caption{Measured latency (s) of single-layer attention with batch size 1 without exponentially decaying causal mask.}
\footnotesize
  \begin{tabular}{|c|c|c|c|c|}
  \hline
  \diagbox{Method}{Seqlen} & 128 & 512 & 2,048 & 8,192 \\ \hline
  \textbf{Vanilla} & $ \textbf{9.56e-5} \pm 3.7e$-$7 $ & $ 3.92e$-$4 \pm 2.7e$-$7 $ & $ 5.91e$-$3\pm 7.7e$-$6 $ & $ 9.4e$-$2 \pm 1e$-$4 $ \\ \hline

\textbf{Bb} & $ 1.06e$-$3 \pm 4.5e$-$5 $ & $ 4.8e$-$3 \pm 5.1e$-$4 $ & $ 1.84e$-$2 \pm 4.7e$-$4 $ & $ 7.08e$-$2 \pm 1.1e$-$3 $ \\ \hline

\textbf{Cdotp} & $ 2.36e$-$4 \pm 2.2e$-$7 $ & $ 9.32e$-$4 \pm 2.7e$-$6 $ & $ 3.7e$-$3 \pm 6.8e$-$6 $ & $ 1.6e$-$2 \pm 2.4e$-$5 $ \\ \hline

\textbf{Cdotp\_torch} & $ 1.43e$-$2  \pm 3.6e$-$5 $ & $ 5.61e$-$2 \pm 3.5e$-$5 $ & $ 2.25e$-$1 \pm 8.6e$-$5 $ & $ 1.21 \pm 4.1e$-$4 $ \\ \hline

\textbf{FA} & $ 1.26e$-$3 \pm 8.2e$-$6 $ & $ 5.46e$-$3 \pm 1.9e$-$5 $ & $ 2.72e$-$2 \pm 1.2e$-$5 $ & $ 1.13e$-$1 \pm 1e$-$4 $ \\ \hline

\textbf{FA\_torch} & $ 6.88e$-$3 \pm 3.6e$-$5 $ & $ 3.3e$-$2 \pm 3.2$-$5 $ & $ 1.29e$-$1 \pm 4.2e$-$5 $ & $ 5.14e$-$1 \pm 8.2e$-$5 $ \\ \hline

\textbf{LA} & $ 1.18e$-$4 \pm 1.3e$-$5 $ & $ \textbf{1.94e-4} \pm 4.8e$-$6 $ & $ \textbf{7.25e-4} \pm 1.7e$-$6 $ & $ \textbf{2.83e-3} \pm 7.4e$-$6 $ \\ \hline

\textbf{LA\_torch} & $ 1e$-$3 \pm 9.7e$-$6 $ & $ 1.35e$-$3 \pm 8e$-$6 $ & $ 2.78e$-$3 \pm 3.2e$-$5 $ & $ 8.84e$-$3 \pm 5.2e$-$5 $ \\ \hline

\textbf{Recur} & $ 6.44e$-$4 \pm 7e$-$7$ & $ 2.8e$-$3 \pm 5.7e$-$6 $ & $ 1.15e$-$2 \pm 1.5e$-$3 $ & $4.57e$-$2 \pm 6.1e$-$5 $ \\ \hline
\end{tabular}

\vspace{2em}

\begin{tabular}{|c|c|c|c|}
  \hline
  \diagbox{Method}{Seqlen} & 12,800 & 25,600 & 100,000 \\ \hline
  \textbf{Vanilla} & \textcolor{red}{OOM} & \textcolor{red}{OOM} & \textcolor{red}{OOM} \\ \hline
\textbf{Bb}  & $ 1.15e$-$1 \pm 1.8e$-$3 $ & $ 2.11e$-$1 \pm 1.9e$-$3 $ & $ 8.9e$-$1 \pm 2.4e$-$2 $ \\ \hline
\textbf{Cdotp} & $ 2.69e$-$2 \pm 5e$-$5 $ & $ 1.07e$-$1 \pm 7.6e$-$5 $ & $ 5.66e$-$1 \pm 7e$-$3 $  \\ \hline
\textbf{Cdotp\_torch}  & $ 1.39 \pm 5.5e$-$4 $ & $ 2.9 \pm 8.2e$-$2 $ & $ 1.11e1 \pm 8.9e$-$2 $  \\  \hline
\textbf{FA} & $ 1.77e$-$1 \pm 1.5e$-$4 $ & $ 3.58e$-$1 \pm 4.3e$-$4 $ & $ 1.41 \pm 1.7e$-$3 $  \\ \hline
\textbf{FA\_torch} & $ 8.03e$-$1 \pm 6.7e$-$6 $ & $ 1.6 \pm 1.3e$-$5 $ & $ 6.25 \pm 1.7e$-$3 $  \\ \hline
\textbf{LA} & $ \textbf{4.45e-3} \pm 5.4e$-$6 $ & $ \textbf{8.69e-3} \pm 2e$-$5 $ & $ \textbf{4.31e-2} \pm 3.2e$-$3 $  \\ \hline
\textbf{LA\_torch} & $ 1.35e$-$2 \pm 2.1e$-$4 $ & $ 2.65e$-$2 \pm 3.7e$-$4 $ & $ 9.96e$-$2 \pm 2.6e$-$4 $  \\ \hline
\textbf{Recur} & $ 9.12e$-$2 \pm 4.6e$-$5 $ & $ 1.82e$-$1 \pm 1.2e$-$4 $ & $ 7.35e$-$1 \pm 1e$-$3 $ \\ \hline
  \end{tabular}
\label{tab:single_bs_1_wo_decay}
\end{table}

\begin{table}[h!]
\centering

\caption{Measured latency (s) of single-layer attention with batch size 1 and exponentially decaying causal mask.}
\footnotesize
\begin{tabular}{|c|c|c|c|c|}
  \hline
  \diagbox{Method}{Seqlen} & 128 & 512 & 2,048 & 8,192 \\ \hline
  \textbf{Vanilla} & $ 1.12e$-$1 \pm 1.7e$-$3 $ & $ 5.08e$-$1 \pm 1.7e$-$3 $ & $ 2.593 \pm 8.2e$-$3 $ & $ 1.86e1 \pm 1.5e$-$1 $     \\ \hline
  \textbf{Bb} & $ 5.53e$-$1 \pm 6.4e$-$2 $ & $ 3.25e$-$1 \pm 4.8e$-$2 $ & $ 5.4e$-$1 \pm 5.1e$-$2 $ & $ 5.56e$-$1 \pm 2.8e$-$2 $   \\ \hline
  \textbf{Cdotp} & $ 2.36e$-$4 \pm 1.3e$-$7 $ & $ 9.28e$-$4 \pm 2.9e$-$6 $ & $ 3.68e$-$3 \pm 7.5e$-$6 $ & $ 1.66e$-$2 \pm 2.3e$-$5 $  \\ \hline
  \textbf{Cdotp\_torch} & $ 1.59e$-$2 \pm 3.2e$-$5 $ & $ 8.37e$-$2 \pm 5.5e$-$5 $ & $ 2.51e$-$1 \pm 5.7e$-$5 $ & $ 1.36 \pm 4.2e$-$4 $ \\ \hline
  \textbf{FA} & $ 1.26e$-$3 \pm 8.2e$-$6 $ & $ 5.49e$-$3 \pm 2.7e$-$5 $ & $ 2.71e$-$2 \pm 1.4e$-$5 $ & $ 1.14e$-$1 \pm 1.2e$-$4 $  \\ \hline
  \textbf{FA\_torch} & $ 1.58e$-$2 \pm 7.3e$-$6 $ & $ 5.57e$-$2 \pm 1.8e$-$5 $ & $ 2.12e$-$1 \pm 3.5e$-$6 $ & $ 8.37e$-$1 \pm 1.4e$-$4 $  \\ \hline
  \textbf{LA} & $ \textbf{1.27e-4} \pm 1.3e$-$5 $ & $ \textbf{1.96e-4} \pm 8e$-$7 $ & $ \textbf{7.49e-4} \pm 2.8e$-$6 $ & $ \textbf{2.9e-3} \pm 1.2e$-$5 $  \\ \hline
  \textbf{LA\_torch} & $ 1.21e$-$3 \pm 6.5e$-$6 $ & $ 1.56e$-$3 \pm 5e$-$6 $ & $ 3.45e$-$3 \pm 8.6e$-$5 $ & $ 1.18e$-$2 \pm 4.2e$-$4 $  \\ \hline
  \textbf{Recur} & $ 2e$-$1 \pm 2.6e$-$3 $ & $ 8.1e$-$1 \pm 6.4e$-$3 $ & $ 3.17 \pm 2e$-$2 $ & $ 1.18e1 \pm 9.7e$-$2 $ \\ \hline
  \end{tabular}

\vspace{2em}

\begin{tabular}{|c|c|c|c|}
  \hline
  \diagbox{Method}{Seqlen} & 12,800 & 25,600 & 100,000 \\ \hline
   \textbf{Vanilla} &  \textcolor{red}{OOM} & \textcolor{red}{OOM} & \textcolor{red}{OOM} \\ \hline
  \textbf{Bb} &  $ 8.53e$-$1 \pm 3.3e$-$2 $ & $ 9.26e$-$1 \pm 4.2e$-$2 $ & $ 1.69 \pm 3.7e$-$2 $  \\ \hline
   \textbf{Cdotp} &  $ 2.7e$-$2 \pm 7.7e$-$5 $ & $ 1.06e$-$1 \pm 6e$-$5 $ & $ 5.64e$-$1 \pm 7.4e$-$3 $  \\ \hline

  \textbf{Cdotp\_torch} &  $ 1.54 \pm 8.6e$-$4 $ & $ 3.46 \pm 1.2e$-$1 $ & $ 1.26e1 \pm 1.6e$-$1 $  \\ \hline

  \textbf{FA} &  $ 1.78e$-$1 \pm 2e$-$4 $ & $ 3.59e$-$1 \pm 5.5e$-$4 $ & $ 1.42 \pm 1.9e$-$3 $  \\ \hline

  \textbf{FA\_torch} &  $ 1.3 \pm 2.1e$-$5 $ & $ 2.6 \pm 3.5e$-$5 $ & $ 1.01e1 \pm 1.7e$-$4 $  \\ \hline

  \textbf{LA} & $ \textbf{4.51e-3} \pm 1.4e$-$5 $ & $ \textbf{9.08e-3} \pm 1.5e$-$5 $ & $ \textbf{4.14e-2} \pm 2e$-$3 $ \\ \hline

  \textbf{LA\_torch} &  $ 4.66e$-$3 \pm 2.5e$-$6 $ & $ 3.21e$-$2 \pm 4.1e$-$4 $ & $ 1.54e$-$1 \pm 3.7e$-$3 $  \\ \hline

  \textbf{Recur} &  $ 2.0e1 \pm 8.6e$-$3 $ & $ 3.41e2 \pm 1.5e2 $ & \textcolor{red}{OOM} \\ \hline

  \end{tabular}

\label{tab:single_bs_1_with_decay}
\end{table}

\begin{table}[h!]
\centering
\footnotesize
\caption{Measured latency (s) of single-layer attention with batch size 16 without exponentially decaying causal mask.}
  \begin{tabular}{|c|c|c|c|c|}
  \hline
  \diagbox{Method}{Seqlen} & 128 & 512 & 2,048 & 8,192 \\ \hline
  \textbf{Vanilla} & $ 5.46e$-$4 \pm 3.4e$-$7 $ & $ 5.33e$-$3 \pm 1.2e$-$5 $ & $ 9.65e$-$2 \pm 2e$-$4 $ & \textcolor{red}{OOM} \\ \hline

\textbf{Bb} & $ 2.54e$-$3 \pm 1e$-$4 $ & $ 9.33e$-$3 \pm 1.1e$-$4 $ & $ 3.68e$-$2 \pm 8.6e$-$5 $ & $ 1.46e$-$1 \pm 4.6e$-$5 $ \\ \hline

\textbf{Cdotp} & $ \textbf{4.42e-4} \pm 1.5e$-$7 $ & $ \textbf{1.82e-3} \pm 7.9e$-$6 $ & $ \textbf{7.5e-3} \pm 2.4e$-$5 $ & $ \textbf{3.15e-2} \pm 6.6e$-$5 $ \\ \hline

\textbf{Cdotp\_torch} & $ 3.31e$-$2 \pm 2.9e$-$6 $ & $ 1.32e$-$1 \pm 1.6e$-$5 $ & $ 5.31e$-$1 \pm 4.5e$-$5 $ & $ 2.12 \pm 1e$-$4 $ \\ \hline

\textbf{FA} & $ 1.85e$-$2 \pm 4.9e$-$5 $ & $ 7.42e$-$2 \pm 1.5e$-$4 $ & $ 3.92e$-$1 \pm 1.4e$-$3 $ & $ 1.48 \pm 1.8e$-$3 $ \\ \hline

\textbf{FA\_torch} & $ 5.9e$-$2 \pm 5.8e$-$6 $ & $ 2.31e$-$1 \pm 3.7e$-$6 $ & $ 9.19e$-$1 \pm 8.2e$-$6 $ & $ 3.67 \pm 3.2e$-$5 $ \\ \hline

\textbf{LA} & $ 5.72e$-$4 \pm 6.6e$-$7 $ & $ 2.33e$-$3 \pm 8.6e$-$6 $ & $ 9.73e$-$3 \pm 3.4e$-$5 $ & $ 4e$-$2 \pm 1.2e$-$4 $ \\ \hline

\textbf{LA\_torch} & $ 1.18e$-$2 \pm 2.3e$-$4 $ & $ 1.82e$-$2 \pm 3.2e$-$5 $ & $ 3.9-2 \pm 4.3e$-$5 $ & $ 1.18e$-$1 \pm 1.9e$-$4 $ \\ \hline

\textbf{Recur} & $ 1.49e$-$3 \pm 6.8e$-$5 $ & $ 8.6e$-$3 \pm 9.3e$-$5 $ & $ 4.36e$-$2 \pm 4.4e$-$5 $ & $ 2.12e$-$1 \pm 1e$-$4 $ \\ \hline

  \end{tabular}

\vspace{2em}

  \begin{tabular}{|c|c|c|c|}
  \hline
  \diagbox{Method}{Seqlen} & 12,800 & 25,600 & 100,000 \\ \hline
  \textbf{Vanilla}  & \textcolor{red}{OOM} & \textcolor{red}{OOM} & \textcolor{red}{OOM} \\ \hline
\textbf{Bb} & $ 2.31e$-$1 \pm 1e$-$4 $ & $ 4.65e$-$1 \pm 8.9e$-$5 $ & \textcolor{red}{OOM} \\ \hline
\textbf{Cdotp} & $ \textbf{4.71e-2} \pm 2.1e$-$5 $ & $ \textbf{9.97e-2} \pm 3.1e$-$4 $ & \textcolor{red}{OOM} \\ \hline
\textbf{Cdotp\_torch}  & $ 3.38 \pm 2.1e$-$4 $ & $ 7.04 \pm 4.5e$-$4 $ & \textcolor{red}{OOM} \\ \hline
\textbf{FA} & $ 2.31 \pm 1.1e$-$3 $ & $ 4.63 \pm 1.6e$-$3 $ & \textcolor{red}{OOM} \\ \hline
\textbf{FA\_torch} & $ 5.75 \pm 5.9e$-$5 $ & $ 1.17e1 \pm 3.4e$-$5 $ & \textcolor{red}{OOM} \\ \hline
\textbf{LA} & $ 6.07e$-$2 \pm 3.5e$-$5 $ & $ 1.27e$-$1 \pm 3.9e$-$4 $  & \textcolor{red}{OOM} \\ \hline
\textbf{LA\_torch} & $ 1.78e$-$1 \pm 2.6e$-$4 $ & $ 3.47e$-$1 \pm 1e$-$4 $ & \textcolor{red}{OOM} \\ \hline
\textbf{Recur} & $3.81e$-$1 \pm 2.8e$-$4 $ & $ 8.2e$-$1 \pm 3.3e$-$4 $  & \textcolor{red}{OOM} \\ \hline
  \end{tabular}
\label{tab:single_bs_16_wo_decay}
\end{table}

\begin{table}[h!]
\centering
\footnotesize
\caption{Measured latency (s) of single-layer attention with batch size 16 and exponentially decaying causal mask.}
  \begin{tabular}{|c|c|c|c|c|}
  \hline
  \diagbox{Method}{Seqlen} & 128 & 512 & 2,048 & 8,192 \\ \hline
  \textbf{Vanilla} &$ 1.12e$-$1 \pm 8.9e$-$4 $ & $ 5.43e$-$1 \pm 1.7e$-$2 $ & $ 2.77 \pm 2.8e$-$2 $ & \textcolor{red}{OOM} \\ \hline
  \textbf{Bb} & $ 4.52e$-$2 \pm 2e$-$3 $ & $ 6.29e$-$2 \pm 7.8e$-$3 $ & $ 8.55e$-$2 \pm 6.2e$-$4 $ & $ 2.93e$-$1 \pm 1.8e$-$2 $ \\ \hline

\textbf{Cdotp} & $ \textbf{4.88e-4} \pm 7.3e$-$7 $ & $ \textbf{2.03e-3} \pm 1.1e$-$5 $ & $ \textbf{8.36e-3} \pm 2.2e$-$5 $ &$ \textbf{3.5e-2} \pm 1.1e$-$4 $ \\ \hline

\textbf{Cdotp\_torch} &  $ 4.58e$-$2 \pm 4.8e$-$6 $ & $ 1.83e$-$1 \pm 2e$-$5 $ & $ 7.34e$-$1 \pm 5.7e$-$5 $ & $ 2.93 \pm 8.5e$-$5 $ \\ \hline

\textbf{FA} & $ 1.85e$-$2 \pm 4.3e$-$5 $ & $ 7.49e$-$2 \pm 8.6e$-$5 $ & $ 3.9e$-$1 \pm 1.1e$-$3 $ & $ 1.48 \pm 1.5e$-$3 $ \\ \hline

\textbf{FA\_torch} &  $ 1.12e$-$1 \pm 1.3e$-$6 $ & $ 4.36e$-$1 \pm 2.1e$-$6 $ & $ 1.74 \pm 2.7e$-$5 $ & $ 6.98 \pm 7.1e$-$5 $ \\ \hline

\textbf{LA} &  $ 6.52e$-$4 \pm 1.5e$-$6 $ & $ 2.57e$-$3 \pm 6.7e$-$6 $ & $ 1.05e$-$2 \pm 2.2e$-$5 $ & $ 4.28e$-$2 \pm 8.6e$-$5 $ \\ \hline

\textbf{LA\_torch} &  $ 1.15e$-$2 \pm 2e$-$4 $ & $ 1.98e$-$2 \pm 2e$-$5 $ & $ 4.49e$-$2 \pm 6.6e$-$5 $ & $ 1.41e$-$1 \pm 2.9e$-$4 $ \\ \hline

\textbf{Recur} &  $ 1.87e$-$1 \pm 2.1e$-$3 $ & $ 7.4e$-$1 \pm 7.7e$-$4 $ & $ 2.98 \pm 4.1e$-$4 $ & $ 1.2e1 \pm 6.7e$-$3 $ \\ \hline

  \end{tabular}

\vspace{2em}

  \begin{tabular}{|c|c|c|c|}
  \hline
  \diagbox{Method}{Seqlen} & 12,800 & 25,600 & 100,000 \\ \hline
  
  \textbf{Vanilla} & \textcolor{red}{OOM} & \textcolor{red}{OOM} & \textcolor{red}{OOM}  \\ \hline
  \textbf{Bb} & $ 3.39e$-$1 \pm 7.2e$-$4 $ & $ 1.31 \pm 3.3e$-$2 $ & \textcolor{red}{OOM} \\ \hline
\textbf{Cdotp} & $ \textbf{5.25e-2} \pm 3.3e$-$5 $ & $ \textbf{1.12e-1} \pm 2.3e$-$4 $ & \textcolor{red}{OOM} \\ \hline
\textbf{Cdotp\_torch} & $ 4.64 \pm 3.2e$-$5 $ & $ 9.59 \pm 4.8e$-$4 $ & \textcolor{red}{OOM} \\ \hline
\textbf{FA} & $ 2.32 \pm 1.2e$-$3 $ & $ 4.65 \pm 1.9e$-$3 $ & \textcolor{red}{OOM} \\ \hline
\textbf{FA\_torch} & $ 1.09e1 \pm 2.2e$-$5 $ & $ 2.165e1 \pm 1.1e$-$4 $ & \textcolor{red}{OOM} \\ \hline
\textbf{LA} & $ 6.49e$-$2 \pm 2.7e$-$5 $ & $ 1.34e$-$1 \pm 2.4e$-$4 $ & \textcolor{red}{OOM} \\ \hline
\textbf{LA\_torch} & $ 2.13e$-$1 \pm 4.2e$-$4 $ & $ 4.16e$-$1 \pm 9.3e$-$5 $ & \textcolor{red}{OOM} \\ \hline
\textbf{Recur} &  $ 2.1e1 \pm 7.6e$-$2 $ & \textcolor{red}{OOM} & \textcolor{red}{OOM} \\ \hline

  \end{tabular}
\label{tab:single_bs_16_with_decay}
\end{table}

\textbf{Result analysis.} 
As a sanity check, we could clearly observe that, as the sequence length of prompts increased, the latency of all the methods (except the \textbf{Vanilla} method) grew linearly. As shown in \Cref{tab:single_bs_1_wo_decay}, when the batch size is 1, the \textbf{Vanilla} method incurs an OOM (out-of-memory) error once the sequence length surpasses 8,192, whereas other linear attention computation methods do not face this issue. 

Additionally, \Cref{tab:single_bs_1_wo_decay,tab:single_bs_1_with_decay} indicate that, with a batch size of 1 and sequence length over 512, lightningAttention-2 (\textbf{LA}) is the fastest, outperforming other methods by at least a factor of two;
while for short inputs (sequence length $\leq 128$), 
the \textbf{Vanilla} method, due to its simplistic design, is the best choice.
In \Cref{tab:single_bs_1_with_decay}, it is also noted that the \textbf{Recur} method leads to an OOM error at a sequence length of 100,000, possibly due to the excessive depth of recursion exceeding the GPU memory limit.

 As shown in \Cref{tab:single_bs_16_wo_decay,tab:single_bs_16_with_decay}, we observe that when the batch size is 16, the \textbf{Cdotp} method outperforms all the other methods across various sequence lengths. 
This superior performance is likely attributed to the CUDA optimizations applied to the batch size dimension. Additionally, for a given sequence length, increasing the batch size from 1 to 16 does not result in a 16$\times$ increase in computational time, as parallel processing occurs along the batch size dimension, thereby enhancing GPU utilization. Furthermore, we observe that the methods implemented with CUDA perform 2$\times$\textasciitilde20$\times$ speedup than those implemented with PyTorch.

\textbf{Suboptimality of the recursive strategy.}
As analyzed in \Cref{sec:recursion}, we observe that when the sequence length is smaller and there is no exponentially decaying causal mask, the performance of \textbf{Recur} is comparable to the best performance; 
the time consumption of recursion increases linearly with the sequence length. 

However, when an exponentially decaying causal mask is used, the performance of \textbf{Recur} is significantly worse than that of other methods. Such a discrepancy is related to the base method it employs (in the original implementation of \textbf{Recur}, when the recursive sequence length falls below 32, the base method is invoked). 
For the cases with the exponentially decaying mask, the computation is performed directly using formulas written in PyTorch, whereas, for the scenario without an exponentially decaying causal mask, the causal-dot-product method is utilized.

\subsection{Effectiveness on linear transformers}
\label{sec:exp-linear-transformer}

We evaluated all the methods in the library w.r.t.\ the improved capability of prompt processing in LLMs.
Experiment settings and result analysis are listed as follows.

\textbf{Large language models.} 
Specifically, we chose two open-source linear transformer large language models, TransNormerLLM-7B, and toy-retnet-1.3B. The two models both support exponentially decaying causal masks. 
More details are provided in~\Cref{sec:experiment_detail}.

\textbf{Datasets and protocols.} 
To prepare longer prompts, we extracted samples from the LongBench~\citep{longbench} datasets. 
Specifically, we set the maximum sequence length to 8k based on our GPU memory size;
i.e., we truncated the prompts with more than 8k tokens and filtered out the prompts shorter than 6k. 

In this experiment, we executed the attention computing in an LLM with various baseline methods under two typical inference settings: the batch size is 1 or 2. 
Each experimental trial was conducted seven times. To reduce the impact of loading phase time variations, we excluded the highest and lowest time measurements and calculated the average of the remaining values to derive the final results. 
More information is in \Cref{sec:experiment_detail}.

\begin{table}[t!]
\centering
\caption{The wall time inference performance of TransNormerLLM-7B uses different methods to go through the dataset when the batch size is 1 and 2. The ``original'' method denotes the original attention computation method implemented in TransNormerLLM-7B. 
}

  \begin{tabular}{|c|c|c|c|c|}
  \hline
  \multirow{2}{*}{\diagbox{Method}{Batch size}} & \multicolumn{2}{c|}{1} & \multicolumn{2}{c|}{2} \\ \cline{2-5}
                          & Avg Time(s) & Std & Avg Time(s) & Std  \\ \hline
  \textbf{Original}        &   64.6       &  0.0031  & 64.6  & 0.0026 \\ \hline
  \textbf{Bb}        &     83.6     & 0.0077   & 73.8  & 0.08 \\ \hline
  \textbf{Cdotp}        &       67.3   & 0.047   & \textbf{ 58.9 } & 0.015 \\ \hline
  \textbf{Cdotp\_torch}        &   268.0      & 1.4   & 159.0  & 0.16 \\ \hline
    \textbf{FA}        &  91.0        & 0.0018   & 88.0  & 0.06 \\ \hline
  \textbf{FA\_torch}        &       308.0   &  0.0047  & 244.0  & 0.0024 \\ \hline
  \textbf{LA}        &    65.4      & 0.017   & 66.1  & 0.0034 \\ \hline
  \textbf{LA\_torch}        &   \textbf{ 63.5}      & 0.027   & 63.6  & 0.035 \\ \hline
  \textbf{Vanilla}        &      3688    & 330   & \textcolor{red}{OOM}  & -\\ \hline
  \textbf{Recur}        &       2539   & 36   & 1402  & 53\\ \hline
  \end{tabular}
\label{tab:transnormerLLM}
\end{table}

\begin{table}[t!]
\centering
\caption{The inference wall time of toy-retnet-1.3B uses different methods to go through the dataset when the batch size is 1 and 2. The \textbf{Original} method denotes the original attention computation method implemented in toy-retnet-1.3B.}
  \begin{tabular}{|c|c|c|c|c|}
  \hline
  \multirow{2}{*}{\diagbox{Method}{Batch size}} & \multicolumn{2}{c|}{1} & \multicolumn{2}{c|}{2} \\ \cline{2-5}
                          & Avg Time(s) & Std & Avg Time(s) & Std  \\ \hline
  \textbf{Original}        &       47.4   & 0.25   & 40.7  & 0.15 \\ \hline
  \textbf{Bb}        &     77.9     & 0.62    & 53.7  & 0.17\\ \hline
  \textbf{Cdotp}        &    39.6      & 0.11   & 36.4  & 0.075 \\ \hline
  \textbf{Cdotp\_torch}        &      299.0    & 1.1   &  179.0 & 0.76 \\ \hline
    \textbf{FA}        &    64.8      & 0.068   & 62.6  & 0.045 \\ \hline
  \textbf{FA\_torch}        &     374.0     & 0.29   & 285.0  & 0.18 \\ \hline
  \textbf{LA}        &     38.0     & 0.12   & 34.0  & 0.12\\ \hline
  \textbf{LA\_torch}        &    \textbf{  37.5}    & 0.081    & \textbf{30.6 } & 0.14\\ \hline
  \textbf{Vanilla}        &  951.0        & 5   & 475.0  & 4.2 \\ \hline
  \textbf{Recur}        &       895.0   & 1.4   & 503.0  & 1.1 \\ \hline
  \end{tabular}

\label{tab:retnet}
\end{table}

\textbf{Result analysis.}
From \Cref{tab:transnormerLLM,tab:retnet}, we can see that these linear attention computation methods significantly reduce GPU memory usage. 
As a clear evidence, when the batch size is set to 2 on TransNormerLLM-7B , these methods will not result in OOM errors. 

We can observe that when the batch size is 2, the latency of each method is mostly lower than that when the batch size is 1; 
this is because when the batch size is 2, only half of the forward propagation is required for all the samples.
In almost all the settings, \textbf{LA\_torch} significantly outperformed the default \textbf{Original} method, which indicates the overhead of its CUDA version \textbf{LA} is non-negligible in these practical settings.
Notably, \textbf{Cdotp}, implemented with the native CUDA, performed the best on TransNormerLLM-7B when the batch size is not 1, aligning with the observation in \Cref{sec:single}.

\section{Conclusions}
\label{sec:conclude}

In this paper, we develop a software tool, \algoname, to efficiently compute exponentially decaying causal linear attention in a single module or an existing linear transformer.
We collect a bundle of computation methods from different application fields, most of which were not originally devoted to the computation of causal linear attention.
We carefully analyze the computational complexity of those algorithms, provide the proof thereof as a reference in this paper, and propose/implement a new method FleetAttention in this package.
Furthermore, we make non-trivial efforts to implement some applicable methods in GPU programming languages such as Triton or CUDA for notably faster performance compared to their native PyTorch counterparts; 
the implementation itself requires deep understanding of GPU architecture, proficient management of parallelization, and usage of numerous intricate techniques, while in developing applications with \algoname~all of those details are hidden from users.

Our implementation in \algoname~is particularly useful when addressing long sequence inputs, which motivates advancements in linear attention and facilitates more efficient processing of long texts in various business fields.
Under various practical scenarios, each method exhibits distinct advantages due to the differences in computational processes; 
a universal subroutine, that selects the (empirically) optimal approach based on the input provided, is integrated in \algoname~as well.

Finally, we wish that this work will inspire the appearance of other open-sourced linear transformers that provides efficient language model services for data science applications in business analytics that can benefit from large language models.

\subsubsection*{Acknowledgments}

We appreciate the valuable advice from Yuzhen Mao, Binhang Yuan, Sitian Chen, Amelie Chi Zhou, Weijia Chen, and Yihu Xu.

\bibliography{ref}
\bibliographystyle{mldraft}
\clearpage
\appendix

\bigskip
\begin{center}
{\large\bf Appendices to ``\mytitle''}
\end{center}

\section{More on Experiments}

\subsection{Evaluation on linear transformers}
\label{sec:experiment_detail}

In \Cref{sec:exp-linear-transformer}, we used TransNormerLLM-7B and toy-retnet-1.3B as the base model to test the implemented methods, on a dataset adapted from the LongBench dataset. 

Since the NVIDIA A6000 GPU used in our experiment solely has 48G memory, the model cannot handle particularly long prompts.
In order to allow the model to handle longer prompts, \ding{182}~we retained the first two layers of the models for inference, so that more memory in the GPU could be used for calculations rather than storage of model weight parameters.
\ding{183}~Moreover, we truncated the prompts in LongBench to 8k tokens using the tokenizer of TransNormerLLM-7B. 
At the same time, we removed questions with length less than 6k.
Consequently, we get an crafted evaluation dataset with 100 prompt samples. 

In addition, as toy-retnet-1.3B only supports the fp32 data type, fp32 numerical precision is adopted along the experiments.
To alleviate the impacts of the GPU startup time in benchmarking computation methods, we conducted $7$ experiments in succession; 
we next removed the maximum and minimum results, and then computed the reported statistics on the remaining results, such as the average value and the sample standard deviation.

\subsection{License information}

TransNormerLLM-7B model is licensed under the Apache 2.0 License.

toy-retnet-1.3B model is licensed under the Apache 2.0 License.

LongBench is licensed under the MIT License. 

\section{Algorithm Details}
\label{sec:other}

\subsection{The ``row-based'' algorithm behind \texttt{causal-dot-product\_torch}}
\label{al:performer}

\citet[Equation~(11)]{choromanski2020rethinking} introduced a prefix sum algorithm for causal linear attention. 
In each iteration, it returns one row of the output matrix, and we thus refer to this algorithm as the ``row-based algorithm.

Vectors $\mtx c_j, \mtx b_j \in \mathbb{R}^{1 \times r}$, and $\mtx v_j \in \mathbb{R}^{1 \times d}$ represent the $j$-th row-vectors in matrices $\mtx{C}, \mtx{B},\mtx{V}$, respectively.
A single row of the output matrix $\mtx O$ is then computed as:
\begin{align}
\label{equ:performer1}
\mtx{O}_i=\sum_{j=1}^i \mtx b_i \mtx c_j^T \mtx v_j = \mtx b_i\sum_{j=1}^i\mtx c_j^T \mtx v_j \in \mb R^{1 \times d}
\end{align}
Let $\mtx{U}_i = \sum_{j=1}^i\mtx c_j^T\mtx v_j \in \mathbb{R}^{r \times d}$, then \Cref{equ:performer1} can be realized through  recursion:
\begin{align}
\label{equ:performer2}
\mtx{O}_i = \mtx b_i \mtx{U}_i, \quad \text{where~} \mtx{U}_i = \mtx{U}_{i-1}+\mtx c_i^T \mtx v_i.
\end{align}
The full row-based algorithm is described in Algorithm~\ref{al:row_based}.
The time complexity for processing each row of $\mtx{O}$ is $\m O(rd)$, leading to a total time complexity of $\m O(Nrd)$. The space complexity is $\m O(Nd+rd)$.

\begin{algorithm}[!t]
  \caption{\small\label{al:row_based} \texttt{Row-based} Algorithm}
  \begin{algorithmic}[1]
    \REQUIRE Matrices $\mtx{B}, \mtx{C} \in \mathbb{R}^{N \times r}$, $\mtx{V} \in \mathbb{R}^{N \times d}$.
    \STATE Initialize $\mtx{U} = (\mtx 0)_{r \times d} \in \mathbb{R}^{r \times d}$, $\mtx{O} = (\mtx 0)_{N \times d} \in \mathbb{R}^{N \times d}$.
    \FOR{$1 \le i \le N$}
      \STATE Extract row vectors $\mtx b_i \in \mathbb{R}^{1 \times r}$, $\mtx c_i \in \mathbb{R}^{1 \times r}$, $\mtx v_i \in \mathbb{R}^{1 \times d}$ from $\mtx{B}, \mtx{C}, \mtx{V}$ respectively.
      \STATE Update $\mtx{U} \leftarrow \mtx{U} + \mtx c_i^T \mtx v_i$.
      \STATE Compute $\mtx o_i = \mtx b_i \mtx{U}$.
      \STATE Store $\mtx{O}_i = \mtx o_i$.
    \ENDFOR
    \STATE Return the output $\mtx{O}$.
  \end{algorithmic}
\end{algorithm}

\subsection{\texttt{Block-based} Algorithm}
\label{al:vq}
The row-by-row computation algorithm described in \Cref{equ:performer2} is considered to be inefficient. Transformer-VQ~\citep[Equation~(28)]{lingle2023transformer} proposed a block-by-block computation method 
designed for its specific low-rank matrix operations. 
We extend this method to a general low-rank matrix product $\tilde{\mtx A}=\mtx{B}\mtx{C}$.

Let the block size be denoted as $B_c$. The block slice $[i B_c:(i+1) B_c]$ is abbreviated as $[i]$. Let $\hat{\mtx K} \in \mathbb{R}^{N \times k}$ be the approximation of $\mtx K$ obtained through the low-rank method in Transformer-VQ or other algorithms. 
Let $\mtx{M} \in \{0,1\}^{B_c \times B_c}$ be a lower triangular causal mask, where $\mtx{M}_{ij}=1$ if $j \leq i$ and $\mtx{M}_{ij}=0$ otherwise.
Then the general computation method can present as follows:
\begin{align}
\mtx O_{[i]} &= (\exp(\mtx Q_{[i]}\hat{\mtx K}_{[i]}^T)\odot \mtx M)V_{[i]}+\sum_{j=1}^{i}\exp(\mtx Q_{[i]}\hat{\mtx K}_{[j]}^T)\mtx V_j\\
&= (\mtx B_{[i]}\mtx C_{[i]}^T \odot \mtx M) \mtx V_{[i]}+\sum_{j=1}^{i}\mtx B_{[i]}\mtx C_{[j]}^TV_j\\
&= (\mtx B_{[i]}\mtx C_{[i]}^T \odot \mtx M) \mtx V_{[i]}+\mtx B_{[i]}\sum_{j=1}^{i}\mtx C_{[j]}^T\mtx V_j
\end{align}
Let $\mtx U_{[i]} = \sum_{j=1}^{i}\mtx C_{[j]}^T\mtx V_{[j]} \in \mathbb{R}^{d \times d}$. Then, we obtain:
\begin{align}
\mtx O_{[i]} &= (\mtx B_{[i]}\mtx C_{[i]}^T \odot \mtx M)\mtx V_{[i]} + \mtx B_{[i]}\mtx U_{[i-1]}\\
\mtx U_{[i]} &= \mtx U_{[i-1]}+\mtx C_{[i]}^T\mtx V_{[i]}
\end{align}
The full block-based algorithm is described in Algorithm~\ref{al:block_based}.
The time complexity for processing each block $\mtx O_{[i]}$ is $\m O(B_c^2r+B_c^2d+B_crd)$; 
there are $\left\lceil\frac{N}{B_c} \right\rceil$ blocks, leading to a total time complexity of $\m O(NB_cr+NB_cd+Nrd)$. 
If we set $B_c = \m O(d)$, the time and space complexity are $\m O(Nrd)$ and $\m O(Nd+rd)$, respectively.

\begin{algorithm}[!t]
  \caption{\small\label{al:block_based} \texttt{block-based} Algorithm}
  \begin{algorithmic}[1]
    \REQUIRE Matrices $\mtx{B}, \mtx{C} \in \mathbb{R}^{N \times r}$, $\mtx{V} \in \mathbb{R}^{N \times d}$, block size $B_c$.
    \STATE Divide $\mtx{V}$ into $T_c = \left\lceil\frac{N}{B_c} \right\rceil$ blocks $\mtx{V}_{[1]}, \dots, \mtx{V}_{[T_c]}$ of size $B_c \times d$ each.
    
    \STATE Divide $\mtx{B}, \mtx{C}$ into $T_c$ blocks $\mtx{B}_{[1]}, \dots, \mtx{B}_{[T_c]}; \mtx{C}_{[1]}, \dots, \mtx{C}_{[T_c]};$ of size $B_c \times r$ each.
    
    \STATE Initialize $\mtx{U} = (\mtx 0)_{r \times d} \in \mathbb{R}^{r \times d}$,  $\mtx{O} = (\mtx 0)_{N \times d} \in \mathbb{R}^{N \times d}$.
    \STATE Let $\mtx{M} \in \{0,1\}^{B_c \times B_c}$ be a lower triangular causal mask.
   
    \FOR{$0 \le i < T_c$}
      \STATE Compute $\mtx o = (\mtx B_{[i]}\mtx C_{[i]}^T \odot \mtx{M})\mtx V_{[i]} + \mtx B_{[i]}\mtx{U}$.
      \STATE Update $\mtx{U} \leftarrow \mtx{U} + \mtx C_{[i]}^T\mtx V_{[i]}$.
      \STATE Compute and store $\mtx O_{[i]} = \mtx o$.
    \ENDFOR
    \STATE Return the output $\mtx{O}$.
  \end{algorithmic}
\end{algorithm}

\subsection{\texttt{Recursion} Algorithm}\label{al:recur}

HyperAttention~\citep[Algorithm~(4)]{han2023hyperattention} proposes a recursive method to compute causal attention. 
This approach starts with partitioning the matrices into two distinct parts: one is masked and the other remains unmasked. 
Following this initial division, the method further recursively partitions the masked part while concurrently applying computation method to the unmasked areas. 

\begin{figure}[t]
    \centering
    \includegraphics[width=0.95\textwidth]{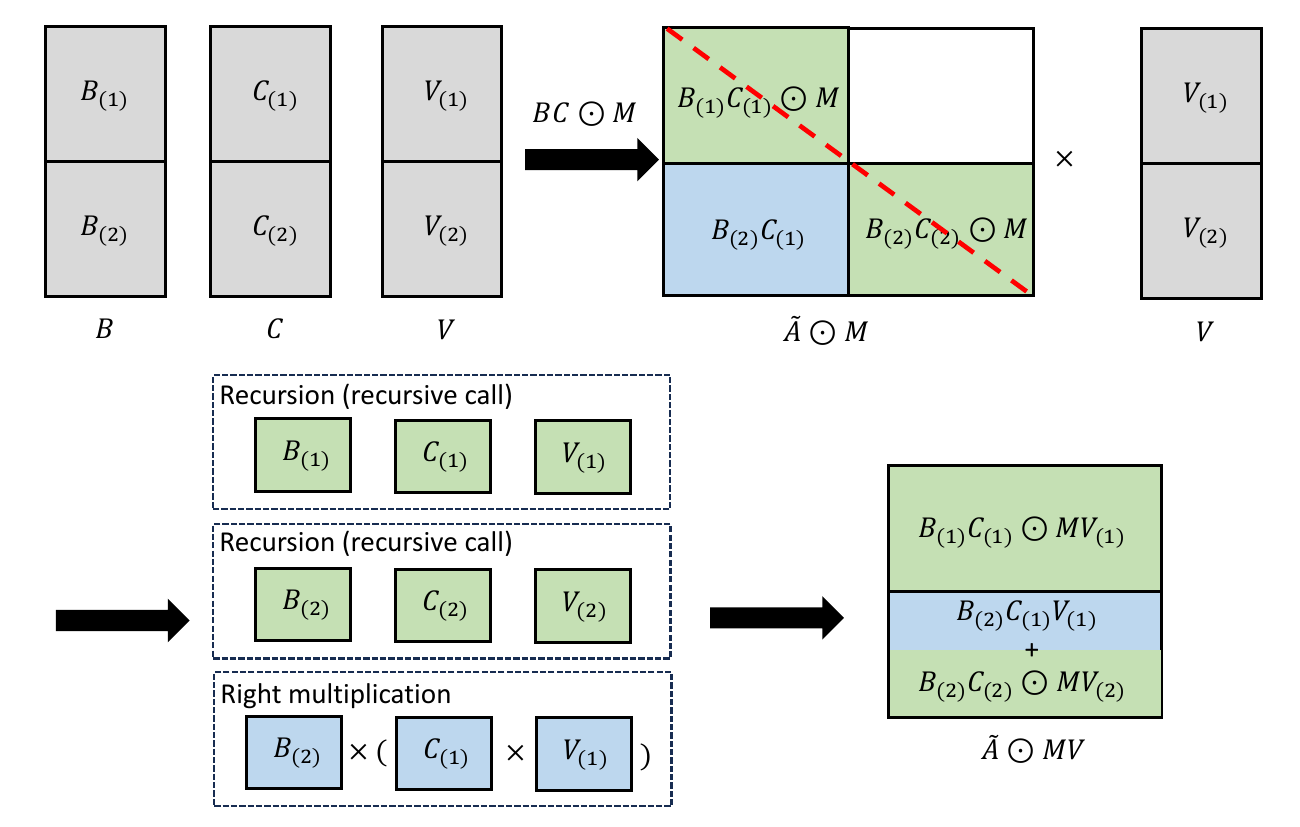}
    \caption{Visualization of the \texttt{Recursion} algorithm. Causal attention matrix can be divided into three equal-sized non-zero sections: $\mtx B_{(1)}\mtx C_{(1)}\odot \mtx M$ and $\mtx B_{(2)}\mtx C_{(2)} \odot \mtx M$ are both masked attention score matrices and $\mtx B_{(2)} \mtx C_{(1)}$ is an unmasked attention score matrix.}
    \label{fig:hpyer} 
\end{figure}

We have extended the recursive approach to the general low-rank matrix form in \Cref{eq:BCMV}. We divide $\mtx B, \mtx C, \mtx V$ recursively to enable more efficient computation of the causal attention.
As shown in Figure~\ref{fig:hpyer}, the masked attention matrix $\tilde{\mtx A} \odot \mtx M$ can be decomposed into three distinct non-zero submatrices, each half the size of the original attention matrix. Notably, the block $\mtx B_{(2)}\mtx C_{(1)}$, located entirely below the diagonal is unmasked attention. We can efficiently compute this part of the result as $\mtx B_{(2)}(\mtx C_{(1)}\mtx V_{(1)})$, enabling linear computation through right multiplication. The two diagonal blocks $\mtx B_{(1)}\mtx C_{(1)}\odot \mtx M$ and $\mtx B_{(2)}\mtx C_{(2)} \odot \mtx M$ in Fig.~\ref{fig:hpyer} represent causal attentions with half the original size. To compute these, we recursively partition them into even smaller blocks and repeat such a decomposition procedure. The full recursion algorithm is described in 
Algorithm~\ref{al:recursion_0} and Algorithm~\ref{al:recursion}.

\begin{algorithm*}[!t]
  \caption{\small\label{al:recursion_0} \texttt{recursion} Algorithm}
  \begin{algorithmic}[1]
    \REQUIRE  Matrices $\mtx{B}, \mtx{C}, \mtx{V}$.
    \IF{$\mathrm{len}(\mtx{B}) \leq \mathrm{termination~block~size}$}
      \STATE Return $(\mtx{M} \odot \mtx{B}\mtx{C}^T) \mtx{V}$
    \ELSE
        \STATE Split $\mtx{B}$, $\mtx{C}$, $\mtx{V}$ into equal sized sub-matrices: $\mtx{B}_1,\mtx{B}_2$; $\mtx{C}_1,\mtx{C}_2$; $\mtx{V}_1,\mtx{V}_2$.
        \STATE $\mtx{O}_1 \leftarrow \mathrm{Recursion}(\mtx{B}_1, \mtx{C}_1, \mtx{V}_1)$
      \STATE $\mtx{O}_2 \leftarrow \mathrm{Recursion}(\mtx{B}_2, \mtx{C}_2, \mtx{V}_2)$
      \STATE $\mtx{O}_3 \leftarrow \mtx{B}_2(\mtx{C}_1^T\mtx{V}_1)$
      \STATE Return $\begin{bmatrix}
 \mtx{O}_1 & \mtx 0\\
 \mtx{O}_3 & \mtx{O}_2
\end{bmatrix}$
    \ENDIF
  \end{algorithmic}
\end{algorithm*}

\begin{algorithm*}[!t]
  \caption{\small\label{al:recursion} \texttt{recursion} Algorithm with special mask}
  \begin{algorithmic}[1]
    \REQUIRE  Matrices $\mtx{B}, \mtx{C}, \mtx{V}$, $w$, left, right.
    \IF{$\mathrm{len}(\mtx{B}) \leq \mathrm{termination~block~size}$}
        \STATE $\mtx{M}_s$ = GetMask(right-left, $w$)
        \STATE Return $(\mtx{M}_s \odot \mtx{B}\mtx{C}^T) \mtx{V}$
    \ELSE
        \STATE Split $\mtx{B}$, $\mtx{C}$, $\mtx{V}$ into equal sized sub-matrices: $\mtx{B}_1,\mtx{B}_2$; $\mtx{C}_1,\mtx{C}_2$; $\mtx{V}_1,\mtx{V}_2$. mid = $\left \lfloor  \frac{\mathrm{left}+\mathrm{right}}{2} \right \rfloor $
        \STATE $\mtx{O}_1 \leftarrow \mathrm{Recursion}(\mtx{B}_1, \mtx{C}_1, \mtx{V}_1, w, \mathrm{left}, \mathrm{mid})$
      \STATE $\mtx{O}_2 \leftarrow \mathrm{Recursion}(\mtx{B}_2, \mtx{C}_2, \mtx{V}_2, w, \mathrm{mid}, \mathrm{right})$
      \STATE $\mtx{W}_1 = [1, w, w^2, ..., w^{\mathrm{right}-\mathrm{mid}-1}]$, $\mtx{W}_2 = [w^{\mathrm{mid}-\mathrm{left}}, w^{\mathrm{mid}-\mathrm{left}-1},..., w]$
      \STATE $\mtx{O}_3 \leftarrow (\mathrm{diag}(\mtx{W}_1)\mtx{B}_2)((\mathrm{diag}(\mtx{W}_2)\mtx{C}_1)^T\mtx{V}_1)$
      \STATE Return $\begin{bmatrix}
 \mtx{O}_1 & \mtx 0\\
 \mtx{O}_3 & \mtx{O}_2
\end{bmatrix}$
    \ENDIF
  \end{algorithmic}
\end{algorithm*}

\begin{figure}[!h]
    \centering
    \includegraphics[width=0.5\textwidth]{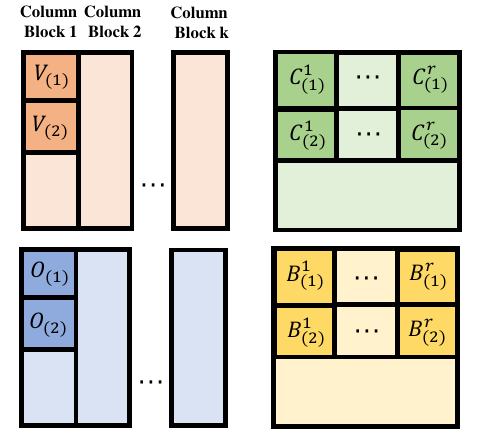}
    \caption{Matrices splitting in FleetAttention }
    \label{fig:FleetAttention_block} 
\end{figure}

\subsection{\texttt{FleetAttention}: An IO-aware Implementation}
\label{app:FleetAttention}

The ideal implementation of FleetAttention on GPU is non-trivial. With vanilla Einstein sum operations, we will need $\m O(Ndr)$ space to store $r$ summands for the final sum up.
It is thus necessary to balance the GPU memory usage and the computation speed, as in regular matrix multiplication on GPU. 
To address this, we introduce an IO-aware implementation,
which maintains the computational complexity at $\m{O}(Ndr)$ while reducing the space complexity to $\m{O}(Nd)$.

FleetAttention leverages the tiling technique~\citep{dao2022flashattention} to enhance memory access efficiency between the GPU High Bandwidth Memory (HBM) and its on-chip Static Random-Access Memory~(SRAM). Meanwhile, this method mitigates memory overflow when dealing with long sequence lengths. The time and memory efficiency of this method are reported in Section~\ref{sec:single}.

The core idea of our optimization is that we split the input matrices $\mtx{B}, \mtx{C}, \mtx{V}$ into blocks. These blocks are then loaded from the slow HBM into the fast SRAM, where the attention outputs are computed w.r.t.\ these blocks. We simultaneously apply the appropriate normalization factors to the outputs of each block, ensuring the accuracy of the results.
Specifically, since the computation for each column of $\mtx{V}$ in~\Cref{equ:p} is independent, we can split $\mtx{V}$ into column blocks for parallel computation. 
Moreover, an entire column block may still exceed the SRAM memory, we further subdivide each column block into row blocks, as illustrated in~\Cref{fig:FleetAttention_block}.
This ensures that the sub-matrices can fit entirely into SRAM, allowing us to complete all operations for a given sub-matrix consecutively.

\begin{algorithm*}[!t]
  \caption{\small\label{al:fleet} \texttt{FleetAttention}}
  \begin{algorithmic}[1]
    \REQUIRE Matrics $\mtx{B}$, $\mtx{C} \in \mathbb{R}^{N \times r}$, $\mtx{V} \in \mathbb{R}^{N \times \boldsymbol{d}}$ in HBM, block sizes $B_c, B_r$.
    \STATE \label{alg:split_v} Divide $\mtx{V}$ into $T_c = \left\lceil\frac{d}{B_c} \right\rceil$ blocks $\mtx{V}^1, \dots, \mtx{V}^{T_c}$ of size $N \times B_c$ each, then devide $\mtx{V}^k$ into $T_r = \left\lceil\frac{N}{B_r} \right\rceil$ blocks $\mtx{V}^{k}_1, \dots, \mtx{V}^{k}_{T_r}$ of size $B_r \times B_c$ each.
    \STATE \label{alg:split_b} Divide $\mtx{B}$ into $T_r \times r$ column blocks $\mtx{B}_1^1, \dots, \mtx{B}_1^{r}, \dots, \mtx{B}_{T_c}^1, \dots, \mtx{B}_{T_r}^r$ of size $B_r \times 1$ each.
    \STATE \label{alg:split_c} Divide $\mtx{C}$ into $T_r \times r$ column blocks $\mtx{C}_1^1, \dots, \mtx{C}_1^{r}, \dots, \mtx{C}_{T_c}^1, \dots, \mtx{C}_{T_r}^r$ of size $B_r \times 1$ each.
    \STATE \label{alg:split_o} Divide the output $\mtx{O}$ into $T_c$ blocks $\mtx{O}^1, \dots, \mtx{O}^{T_c}$ of size $N \times B_c$ each, then devide $\mtx{O}^k$ into $T_r$ blocks $\mtx{O}^{k}_1, \dots, \mtx{O}^{k}_{T_r}$ of size $B_r \times B_c$ each.
    \FOR{$1 \le k \le T_c$, parallel computation of $\mtx{O}^k$} 
    \STATE $\mtx l^0, \dots, \mtx l^r = (\mtx 0)_{B_c} \in \mathbb{R}^{B_c}$
    \FOR{$1 \le i \le T_r$} \label{alg:v_loop}
      \STATE Load $\mtx{V}^{k}_i$ from HBM to on-chip SRAM.
      \STATE On chip, initialize $\mtx{O}_i^k = (\mtx 0)_{B_r \times B_c} \in \mathbb{R}^{B_r \times B_c}$.
      \FOR{$1 \le j \le r$} \label{alg:bc_loop}
        \STATE Load $\mtx{B}^{j}_i, \mtx{C}^{j}_i$ from HBM to on-chip SRAM.
        \STATE On chip, compute $\mtx{O}_i^k \leftarrow \mtx{O}_i^k + \mathrm{diag}(\mtx{B}^{j}_i) (\mathrm{cumsum}(\mathrm{diag}(\mtx{C}^{j}_i) \mtx{V}^{k}_i)+\mtx l^j)$.
        \STATE On chip, compute $\mtx l^j \leftarrow \mathrm{sum}(\mathrm{diag}(\mtx{C}^{j}_i) \mtx{V}^{k}_i) + \mtx l^j $
      \ENDFOR
      \STATE Write $\mtx{O}^{k}_i$ to HBM as the block of $\mtx{O}^{k}_i $. 
    \ENDFOR
    \ENDFOR
\STATE Return the output $\mtx{O}$.
  \end{algorithmic}
\end{algorithm*}

The computation process is described below. To simplify, we focus on a single column block of the value matrix $\mtx{V}$, divided into $\begin{bmatrix} \mtx{V}_{(1)}\\ \mtx{V}_{(2)} \end{bmatrix}$ for some matrices $\mtx{V}_{(1)}$, $\mtx{V}_{(2)} \in \mathbb{R}^{ B_r \times B_c}$, where $B_r$ and $B_c$ are the row and column block sizes, respectively. Our objective is to calculate the attention output $\mtx{O}$ for this column block, which takes a similar form $\begin{bmatrix} \mtx{O}_{(1)}\\ \mtx{O}_{(2)} \end{bmatrix}$ for some metrics $\mtx{O}_{(1)}$, $\mtx{O}_{(2)} \in \mathbb{R}^{ B_r \times B_c}$. 
We have two matrices $\mtx{B}$ and $\mtx{C}$, of the low-rank approximation form 
$$\mtx{C} = \begin{bmatrix}
 \mtx{C}_{(1)}^1, ..., \mtx{C}_{(1)}^r\\
 \mtx{C}_{(2)}^1, ..., \mtx{C}_{(2)}^r
\end{bmatrix}, \qquad
\mtx{B} = \begin{bmatrix} \mtx{B}_{(1)}^1, ..., \mtx{B}_{(1)}^r\\
\mtx{B}_{(2)}^1, ..., \mtx{B}_{(2)}^r \end{bmatrix},$$ where $\mtx{B}_{(1)}^i$, $\mtx{B}_{(2)}^i$, $\mtx{C}_{(1)}^i$, $\mtx{C}_{(2)}^i \in \mathbb{R}^{ B_r \times 1}$ are columns of matrices.

For the first block $\mtx V_{(1)}$, according to \Cref{equ:p}, we can obtain 
\begin{align*}
\mtx O_{(1)} =\sum_{i=1}^r \mathrm{diag}(\mtx{B}_{(1)}^i)\mathrm{cumsum}(\mathrm{diag}(\mtx{C}_{(1)}^i)\mtx{V}_{(1)}),
\end{align*}
where $\mathrm{cumsum}(\cdot)$ returns a matrix of the same shape as the input, representing the cumulative sum of elements along the matrix's vertical direction.
To facilitate the cumsum operation in the subsequent block, we cached the prefix sums of this block:
\begin{align*}
\mtx l_{(1)}^i &= \mathrm{sum}(\mathrm{diag}(\mtx{C}_{(1)}^i)\mtx{V}_{(1)})
\end{align*}
where $\mathrm{sum}(\cdot)$ returns a row vector that is the sum of all the rows in the matrix.
For subsequent blocks, we add prefix sums at the corresponding positions to ensure the accuracy of the results and continue updating the prefix sums. The following recursive formula can be derived:
\begin{align*}
\mtx{O}_{(n)} &=\sum_{i=1}^r \mathrm{diag}(\mtx{B}_{(n)}^i)(\mathrm{cumsum}(\mathrm{diag}(\mtx{C}_{(n)}^i)\mtx{V}_{(n)})+\mtx l_{(n-1)}^i)\\
\mtx l_{(n)}^i &= \mathrm{sum}(\mathrm{diag}(\mtx{C}_{(n)}^i)\mtx{V}_{(n)}) +\mtx l_{(n-1)}^i
\end{align*}
The full procedure of FleetAttention is given in Algorithm~\ref{al:fleet}, and we implemented it in Triton in \algoname.

\subsection{Details about CUDA Implementations}
\label{app:cuda-details}

\texttt{\textbf{causal-dot-product}} natively supported only the 32-bit floating-point (fp32) type and the classic causal linear attention calculations with masks consisting exclusively of 0s and 1s. We have extended this by implementing a half-precision (16-bit floating-point) version using CUDA, enabling its application in popular linear transformer large language models. Additionally, the method now supports exponentially decaying masks, further enhancing its versatility and performance. 

\texttt{\textbf{lightningAttention-2}} is written in OpenAI's Triton programming language, and the official implementation of \texttt{\textbf{lightningAttention-2}} natively supports only 16-bit floating-point computations and is deeply tied to specific GPU; i.e., it could only run on GPUs such as Nvidia A100 or more advanced, otherwise users will encounter resource limitations. 
We thoroughly optimized the implementation to support 32-bit floating-point computations and adjusted parameters for different GPUs, enabling its use for GPUs less advanced than A100. 

Our proposed \texttt{\textbf{FleetAttention}} has two implementations: PyTorch and Triton. 
The PyTorch version is simple to use, and the algorithm is described in \Cref{equ:p}. The Triton version is much more complicated, mainly because it needs to consider parallelization to maximize the utilization of GPU cores. In addition, since the GPU memory is limited, in the implementation we specifically tweak how to set the scale of parallel operations.

\section{Derivations Omitted in the Main Text}

\subsection{Generalization of FleetAttention to Attention with Exponentially Decaying Positional Embedding}
\label{sec:discount_cumsum}

In the specific attention variant with exponentially decaying positional embedding, we have a slightly different mask matrix
$$
\mtx M = [\lambda^{i-j} \cdot \delta(i, j)], \quad \text{with the discount factor~} \lambda \in (0, 1],
$$
which is still a lower triangular matrix.
With the notation above, the original causal attention matrix can be taken as a special case with $\lambda = 1$.

We first study a toy case, in which the $i$-th row of the product $\mtx O = \mtx{MV}$ (i.e., $\mtx B, \mtx C$ is an all-one column vector), and have
\begin{align*}
\mtx O_{i, :} &= \sum_{j=1}^i \lambda^{i-j} \mtx V_{j, :}
= \mtx V_{i, :} + \sum_{j=1}^{i-1} \lambda^{i-j} \mtx V_{j, :}
= \mtx V_{i, :} + \lambda \paren{\sum_{j=1}^{i-1} \lambda^{i-1-j} \mtx V_{j, :}} \\
&= \mtx V_{i, :} + \lambda \mtx O_{i-1, :}.
\end{align*}
We note the operator $\mtx M$ is now equivalent to the so-called discounted $\mathrm{cumsum}$ operator,
which enjoys the linear $\m O(N)$ complexity as well.

Denoting the discounted $\mathrm{cumsum}$ operator as $\lambda$-$\mathrm{cumsum}$,
we combine the pieces above and have
\begin{align*}
\mtx{O} &= \paren{(\mtx{B}\mtx{C}^T)\odot \mtx{M} }\mtx{V}
= \sum_{i=1}^r \mathrm{diag}(\mtx b_i)\mtx{M} \mathrm{diag}(\mtx c_i)\mtx{V} \\
&= \sum_{i=1}^r \mathrm{diag}(\mtx b_i) \cdot \lambda\text{-}\mathrm{cumsum}(\mathrm{diag}(\mtx c_i) \mtx{V}).
\end{align*}
Therefore, the complexity analysis of our algorithm in \Cref{sec:fleetAttention} still applies,
and we conclude the generalized version of FleetAttention maintains $\m O(N)$ time complexity.

\subsection{Proof of \Cref{lem:recur}}
\label{sec:recur}
\begin{proof}
We first call $\mtx{B}$, $\mtx{C} \in \mb{R}^{N \times r}$ , $\mtx{V} \in \mb{R}^{N \times d}$ which satisfies $N\gg d$ and $N\gg r$. $\mtx{M}$ is a $0$-$1$ matrix with elements in the lower triangle all one. $\tilde{\mtx{A'}}\mtx{V} = \mtx{BC} \odot \mtx{M}\mtx{V}$ can be rewritten in a partitioned manner:
\begin{align*}
\begin{pmatrix}
\mtx B_1 \\
\mtx B_2 ,
\end{pmatrix}
\begin{pmatrix}
\mtx C_1, \mtx C_2 
\end{pmatrix} 
\odot 
\mtx{M} 
\begin{pmatrix}
\mtx V_1 \\
\mtx V_2 ,
\end{pmatrix}
= 
\begin{pmatrix}
\mtx B_1 \mtx C_1 \odot \mtx M , \mtx 0 \\
\mtx B_2 \mtx C_1 , \mtx B_2 \mtx C_2 \odot \mtx M
\end{pmatrix}
\begin{pmatrix}
\mtx V_1 \\
\mtx V_2 ,
\end{pmatrix}
= 
\begin{pmatrix}
\mtx B_1 \mtx C_1 \odot \mtx M \mtx V_1  \\
\mtx B_2 \mtx C_1 \mtx V_1 + \mtx B_2 \mtx C_2 \odot \mtx M \mtx V_2
\end{pmatrix}
\end{align*}
For the formula $ \mtx B_1 \mtx C_1 \odot \mtx M \mtx V_1 $ and $ \mtx B_2 \mtx C_2 \odot \mtx M \mtx V_2 $ we can use the method just now to further recurse. For the formula $ \mtx B_2 \mtx C_1 \mtx V_1 $ we can use right multiplication which has a time complexity of $ \m O(N)$. \\
According to the above recursion relationship, we can get the time complexity expression:
\begin{align*}
T(N) &= 2T(N/2) + N \\
\text{Master~theorem}: \quad T(N) &= aT(N/b) + f(N)  \\
\text{So}: \quad \quad \quad a&=2, b=2, f(N)=N.\\
f(N)&=\Theta(N^{log_ba}) =\Theta(N)\\
\text{then}: \quad T(N)&=\Theta(NlgN)
\end{align*}
The claim in \Cref{lem:recur} is then attained.
\end{proof}

\begin{table}[t!]
\centering
\footnotesize
\caption{Runtime analysis of the single-layer attention for both FleetAttention and FleetAttention\_modify methods. The ``bz'' column represents the batch size. The ``exp. decay'' column indicates whether an exponentially decaying causal mask is used.}
  \begin{tabular}{|c|c|c|c|c|c|c|c|c|c|}
  \hline
  \diagbox{Method}{Seqlen} & bz & exp. decay & 128 & 512 & 2048 & 8192 & 12800 & 25600 & 100000 \\ \hline
  \textbf{FA} & 1 & \XSolidBrush  &  1.26e-3 & 5.46e-3  & 2.72e-2 & 1.13e-1 & 1.77e-1  & 3.58e-1 & 1.41 \\ \hline
\textbf{FA} & 1 & \Checkmark  &  1.26e-3 & 5.49e-3  & 2.71e-2 & 1.14e-1 & 1.78e-1 & 3.59e-1 & 1.42 \\ \hline
\textbf{FA} & 16 & \XSolidBrush  &  1.85e-2 & 7.42e-2   & 3.92e-1 & 1.48 & 2.31 & 4.63 & \textcolor{red}{OOM} \\ \hline
\textbf{FA} & 16 & \Checkmark  & 1.85e-2  & 7.49e-2  & 3.90e-1 & 1.48 & 2.32 & 4.65 & \textcolor{red}{OOM} \\ \hline
\textbf{FA\_modify} & 1 & \XSolidBrush  & 7.00e-4  & 3.31e-3 & 1.32e-2 & 5.28e-2 & 8.27e-2 & 1.65e-1 & 6.55e-1 \\ \hline
\textbf{FA\_modify} & 1 &  \Checkmark &  7.17e-4 & 3.33e-3  & 1.33e-2 & 5.43e-2  & 8.34e-2& 1.68e-1 & 6.65e-1 \\ \hline
\textbf{FA\_modify} & 16 &  \XSolidBrush & 1.18e-2 & 5.12e-2  & 2.06e-1 & 8.36e-1  & 1.31 & 2.60& \textcolor{red}{OOM} \\ \hline
\textbf{FA\_modify} & 16 & \Checkmark  & 1.18e-2 & 5.15e-2  & 2.05e-1 & 8.14e-1 & 1.33 & 2.65 & \textcolor{red}{OOM}\\ \hline
  \end{tabular}
\label{tab:single_FleetAttention}
\end{table}

\section{A finer performance analysis of FleetAttention}
\label{sec:analysis_FleetAttention}

The FleetAttention (FA as a shorthand) method performs the sum operations across the rank dimension as an ``atomic operation''.
This ``atomic operation'' necessarily hinders the execution of the FA method,
while we have to set it as atomic to prevent data inconsistency.

To explore the potential benefit of accelerating this ``atomic operation'', we modified its Triton implementation (named FleetAttention\_modify, or FA\_modify in short) so that the sum operations across the rank dimension now occurs directly after loading the data, before writing it back. 
This modification means the operation is no longer atomic, potentially causing the situation where other parallel ranks may load the data (this modification can result in data inconsistency and thus not incorporated in \algoname). 

We have conducted a runtime analysis of both methods following the same experiment setup in \Cref{sec:single}. The results are summarized in \Cref{tab:single_FleetAttention}. 
Under various conditions, where the batch size is either 1 or 16, and the sequence length ranges from 128 to 100,000, the FA\_modify method consistently requires approximately half the time compared to the FA method.
This indicates that parallelizing in the rank dimension indeed affects the performance of the algorithm at the point where results are accumulated.
\end{document}